\documentclass[review]{elsarticle}

\usepackage{times}
\usepackage{soul}
\usepackage{url}
\usepackage{array}
\usepackage[utf8]{inputenc}
\usepackage[small]{caption}
\usepackage{graphicx}
\usepackage{amsmath}
\usepackage{booktabs}
\usepackage{algorithm}
\usepackage{algorithmic}
\usepackage{longtable}
\usepackage[linewidth=1pt]{mdframed}
\usepackage{fancybox,framed}
\usepackage{svg}
\usepackage{multirow}

\usepackage{refcount}

\newcommand{\myfootnotetext}[1]{\footnotetext{#1\label{fn:text}%
        \edef\fnmark{\getpagerefnumber{fn:mark}}%
        \edef\fntext{\getpagerefnumber{fn:text}}%
        \ifx\fnmark\fntext\else\ClassWarning{}{footnote mark and text on different pages!}\fi}}

\usepackage{lineno}
\modulolinenumbers[5]

\journal{Journal}

\biboptions{authoryear}

\begin{document}

\begin{frontmatter}

\title{Deep Patent Landscaping Model \\ Using the Transformer and Graph Embedding}

\author[gachon_address]{Seokkyu Choi}

\author[gachon_address]{Hyeonju Lee}

\author[papago]{Eunjeong Park\corref{mycorrespondingauthor}}
\ead{lucy.park@navercorp.com}

\author[gachon_address]{Sungchul Choi\corref{mycorrespondingauthor}}
\ead{sc82.choi@gachon.ac.kr}

\cortext[mycorrespondingauthor]{Corresponding authors}

\address[gachon_address]{TEAMLAB, Department of Industrial and Management Engineering, Gachon University, \\
Seongnam-si, Gyeonggi-do, Republic of Korea}
\address[papago]{NAVER Corp., Seongnam-si, Gyeonggi-do, Republic of Korea}

\begin{abstract}
  
Patent landscaping is a method used for searching related patents during a research and development (R\&D) project. To avoid the risk of patent infringement and to follow current trends in technology, patent landscaping is a crucial task required during the early stages of an R\&D project. As the process of patent landscaping requires advanced resources and can be tedious, the demand for automated patent landscaping has been gradually increasing. However, a shortage of well-defined benchmark datasets and comparable models makes it difficult to find related research studies.

In this paper, we propose an automated patent landscaping model based on deep learning. To analyze the text of patents, the proposed model uses a modified transformer structure. To analyze the metadata of patents, we propose a graph embedding method that uses a diffusion graph called Diff2Vec. Furthermore, we introduce four benchmark datasets for comparing related research studies in patent landscaping. The datasets are produced by querying Google BigQuery, based on a search formula from a Korean patent attorney. The obtained results indicate that the proposed model and datasets can attain state-of-the-art performance, as compared with current patent landscaping models. 
\end{abstract}


\end{frontmatter}


\section{Introduction}
A patent is a significant deliverable in research and development (R\&D) projects. A patent protects an assignee’s legal rights and also represents current trends in technology. To study technological trends and identify potential patent infringements, most R\&D projects include patent landscaping. Patent landscaping involves collecting and analyzing patent documents related to a specific project~(\cite{bubela2013patent,Wittenburg2015-hv,bubela2013patent,Abood2018-fd}). Generally, patent landscaping is a human-centric, tedious, and expensive process(\cite{trippe2015guidelines,Abood2018-fd}). Researchers and patent attorneys query related patents in large patent databases (by creating keyword candidates), eliminate unrelated patent documents , and extract only valid patent documents related to their project(\cite{Yang2010-mi,Wittenburg2015-hv}). However, as  the participants of the process must be familiar with the scientific and technical domains, these procedures are costly. Furthermore, the patent landscaping task has to be repeated regularly (weekly or monthly) during a project in progress, to search for newly published patents.



In this paper, we propose a supervised deep learning model for patent landscaping. The proposed model aims to eliminate repetitive and inefficient tasks by employing deep learning-based classification models. The proposed model incorporates a modified transformer structure~(\cite{NIPS2017_7181}) and a graph embedding method using a diffusion graphc~(\cite{10.1007/978-3-319-73198-8_9}). As a patent document can contain several textual features and bibliometric data, the modified transformer structure is applied for processing textual data, and the diffusion graph Diff2Vec is applied for processing graph-based bibliometric data fields.


Additionally, as we also aim to contribute resources towards machine learning-based patent landscaping research, we propose benchmark datasets for patent landscaping. Conventionally, owing to issues such as high cost and data security, benchmark datasets for patent landscaping have not been open and available. The proposed benchmark datasets are based on the Korea Intellectual Property Strategy Agency (KISTA\footnote{https://www.kista.re.kr/})’s patent trend report, which was written by human experts, e.g., patent attorneys. We build the benchmark datasets from Google BigQuery by using keyword queries and valid patents from the KISTA patent trends report, as filtered by experts. The experimental results indicate that the proposed model (with the proposed benchmark datasets) outperforms other existing classification models, and the average classification accuracy for each dataset can be improved by approximately 15\%. 


\section{Patent landscaping}

\begin{figure*}[htbp]
    \centering
    \includegraphics[width=\textwidth]{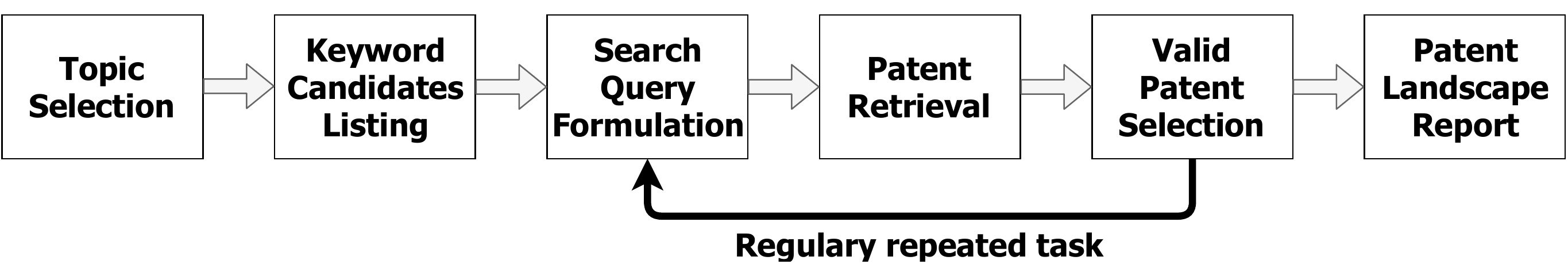}
    \caption{General process of patent landscaping}
    \label{fig:fig_1}
\end{figure*}
 

The entire process of patent landscaping is shown in Figure \ref{fig:fig_1}. First, keyword candidates for the target technology area are extracted to form a search formula or query for patent documents. As many assignees do not allow their patents to be discovered easily to gain an advantage in infringement issues that may arise, they tend to write patent titles and abstracts very generically or to omit technical details(\cite{tseng2007text}). Considering this, a complicated search formula should be created to extract as many relevant patent candidates as possible(\cite{magdy2011study}). The search formula depends on the patent search system that performs the search. For example, a search query for an underwater vehicle device might be created as shown in the box below.

\begin{framed}
((( virtual* or augment* or mixed* ) or ( real* or environment* or space )) or ( augment* and real* )) and ( (( offshore* or off-shore* or ocean ) or ( plant* or platform* )) or ship* or dock* or carrier or vessel or marine or boat* or drillship or ( drill or ship ) or FPSO or ( float* or ( product* or storag* )) or FPU or LNG or FSRU or OSV or aero* or airplane or aircraft or construction or ( civil or engineer* ) or bridge or building or vehicle or vehicular or automotive or as follows automobile )
\end{framed}


As Figure \ref{fig:fig_1} shows, most parts of the process are conducted manually by experts with a technical background, and some parts of the process are repeated. The primary focus of this paper is the regularly repeated task of returning to the search query formulation from a valid patent selection. Once the search formula is created, it is necessary to track new patents (which are regularly published) using a similar search formula. As selecting a first valid patent is similar to creating a training dataset for supervised learning, it can be used to solve repetitive tasks with text classification. As these repetitive tasks require significant unnecessary effort from experts, there is a high possibility of improving them by using a machine learning approach. 


This is not the first study related to the machine learning-based patent landscaping. An automated patent landscaping (APL) approach was previously proposed by~\cite{Abood2018-fd}. They composed a dataset for patent landscaping using “seed patents” created by experts in patent law. Then, they applied a neural network to classify the patents in the collected data as “close to seed patents.” Moreover, they expanded the dataset using related patents. First, they asked experts to designate key patent documents for each technology area as seed patents. Subsequently, they expanded the patent dataset starting with seed patents by using metadata such as cooperative patent classification (CPC) and patent family. 


Although the previous APL study opened the possibility of machine learning-based patent landscaping, there are problems regarding the usage of comparable benchmark datasets. First, there is no suggestion of a comparable set of benchmarking data. There may be situations in which the proposed dataset is generated in a heuristic way, and the learned model learns that heuristic. The dataset is different from a dataset generated by human experts, and it is difficult to generate a model that can replace the intellectual activity of human patent analysis. Moreover, in the APL study, the dataset included patents in very broad and/or common technology fields, such as “machine learning” and “IoT.” However, a typical patent landscaping is conducted on very specific technologies, depending on the projects of companies or research laboratories. We believe these differences make it difficult to apply the previous APL approach to the actual patent landscaping tasks.

In addition to the APL study, some studies on machine learning-based patent classification have suggested models employing the International Patent Code (IPC) classification and long short-term memory and have proposed a model based on a text convolutional neural network (text-CNN)(~\cite{sureka2009semantic,chen2011ipc,lupu2013patent,shalaby2018lstm,doi:10.1162/neco.1997.9.8.1735,li2018deeppatent}). However, the biggest weakness of these studies is the lack of a suitable benchmark dataset, as in the case with APL. Moreover, unlike the actual patent landscaping, IPC classification studies simply predict a fitting IPC code for each patent, which is already granted to all patents by assignees and patent examiners. Thus, these methods are not suitable for patent landscaping in the real world.

\section{KISTA datasets for patent landscaping}

\begin{figure*}[tp]
    \centering
    \includegraphics[width=\textwidth]{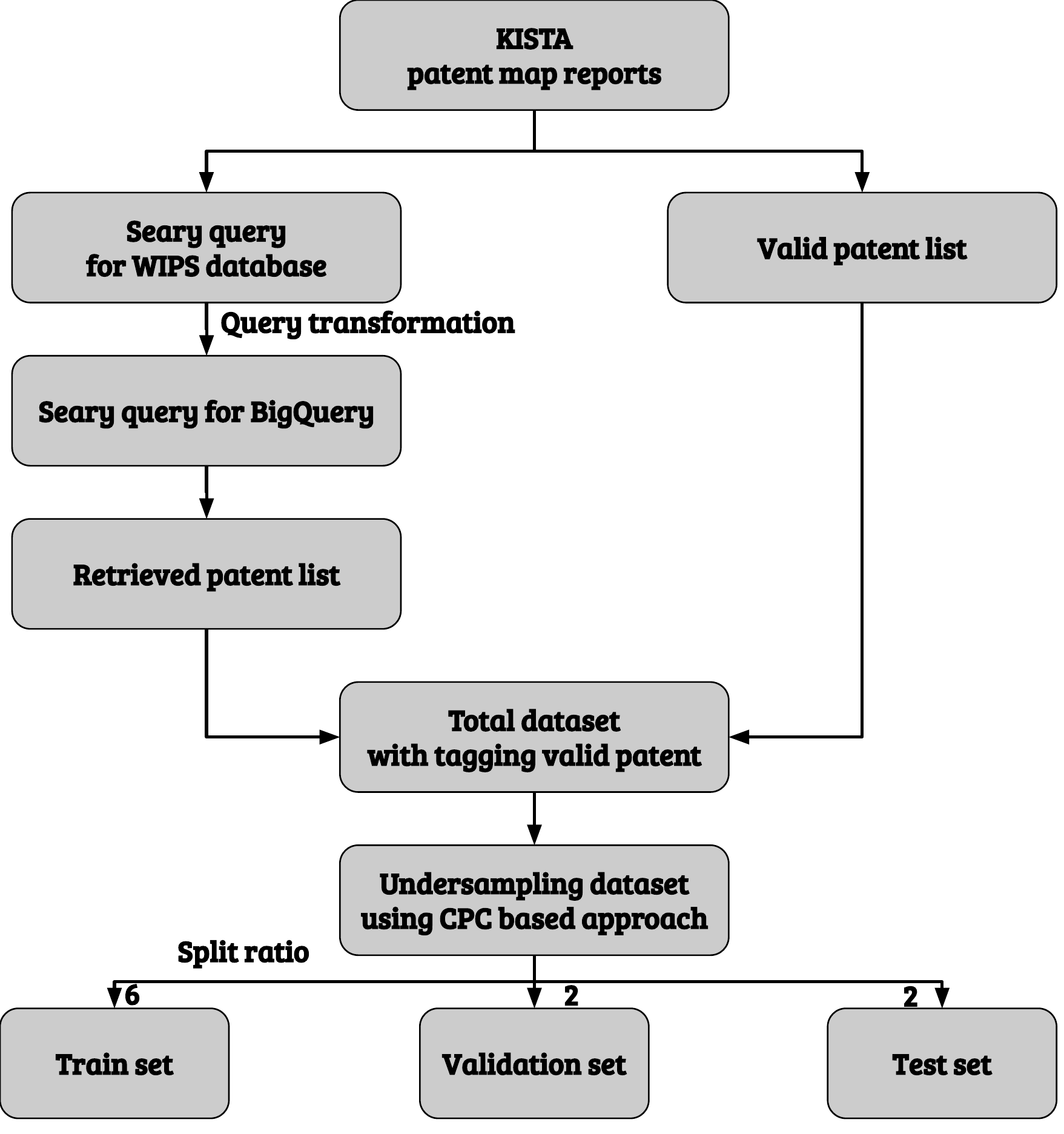}
    \caption{The general process of patent landscaping}
    \label{fig:data_build_process}
\end{figure*}

First, we build datasets using KISTA patent report maps. A detailed flowchart is shown in Figure \ref{fig:data_build_process}.

\subsection{Data sources}
We provide a benchmarking dataset for patent landscaping based on the KISTA patent trends reports\footnote{http://biz.kista.re.kr/patentmap/}. Every year, Korean Intellectual Property Office(KIPO)\footnote{https://www.kipo.go.kr/} publishes more than 100 patent landscaping reports through KISTA. In particular, most reports are available to validate the results of the trends report by disclosing the valid patent list together with the patent search query\footnote{Most of the search queries were based on WIPS (https://www.wipson.com) service, which is a local Korean patent database company.}. Currently, more than 2,500 reports are disclosed. The kinds of technology in the reports are specific, concrete, and sometimes include fusion characteristics. We have constructed datasets for the four technologies listed in Table ~\ref{tab:tab_1}. 

We provide a benchmarking dataset for patent landscaping based on KISTA patent trends reports\footnote{http://biz.kista.re.kr/patentmap/}. Each year, the Korean Intellectual Property Office\footnote{https://www.kipo.go.kr/} publishes more than 100 patent landscaping reports through KISTA. Most trend reports support the findings by disclosing the valid patent list together with the patent search query\footnote{Most of the search queries were based on WIPS (https://www.wipson.com) service, which is a local Korean patent database company.}. Currently, more than 2,500 reports have been disclosed. The types of technology in these reports are specific, concrete, and sometimes include fusion characteristics. We have constructed datasets for the four technologies listed in Table ~\ref{tab:tab_1}.

\begin{table}
\small

    \centering
    \resizebox{\textwidth}{!}{
    \begin{tabular}{c l l}
    \toprule
    Dataset & Full name & Important keywords\\ 
    \midrule
    MPUART & Marine Plant Using Augmented Reality Technology & hmd, photorealistic, georegistered \\ 
    1MWDFS & Technology for 1MW Dual Frequency System & reverse conductive, mini dipole\\ 
    MRRG & Technology for Micro Radar Rain Gauge & klystron, bistatic, frequencyagile\\ 
    GOCS & Technology for Geostationary Orbit Complex Satellite & rover, pgps, pseudolites\\ 
    \bottomrule
    \end{tabular}
    }
    \caption{Patent landscaping benchmarking dataset}
    \label{tab:tab_1}
\end{table}

\subsection{Data acquisition}


To ensure reproducibility in building patent datasets, we built the benchmark datasets using Google BigQuery public datasets. Most of the patent data in the KISTA report were obtained using a search query of a local Korean patent database service called the WIPS. We constructed a Python module for converting the WIPS query into a Google BigQuery service query, extracted the patent dataset from the BigQuery, and marked valid patents among the extracted patents. In a patent search, different datasets could be extracted, depending on the type of publication date and database to be searched. Therefore, we excluded queried patents published after the original publication date depicted in the report. The BigQuery search queries that we used for patent retrieval have been added to Appendix I.

\subsection{Dataset description}

In general, broad and common search keywords are selected for patent retrieval. This is because patent assignees purposely write their patents in plain language, so that competitors cannot find their patents. As a result, patent retrieval by keywords results in a large number of patent documents being searched; unrelated patent documents are excluded from the patent landscaping process by experts.

We searched for the United States Patent and Trademark Office (USPTO) patents in four technology areas, using the above-mentioned search query. As a result, more than a million patent documents were retrieved in three of the four technology domains searched. Among the retrieved patent documents, we designated “valid patents” as those related to the technology areas in the KISTA report. In terms of the classification problems, “valid patent” indicates the “true Y label” to be classified. The number of valid patents is less than 1000 in all domains. Hence, these datasets are imbalanced: most retrieved results are not “valid patents.” We obtain patent information, including metadata from BigQuery, to indicate whether or not they are valid. The final dataset is described in Table ~\ref{tab:tab_2}.

\begin{table}[h]
\centering
\begin{tabular}{cccc} 
\toprule
Dataset name & \# of patents & \# of valid patents & Data URL\\
\midrule
MPUART & 1,469,741 & 468 & https://bit.ly/343JSD8\\
1MWDFS & 1,774,132 & 927 & https://bit.ly/2Wk7kJI\\
MRRG & 2,068,566 & 225 & https://bit.ly/2BTdKGe\\
GOCS & 294,636 & 653 & https://bit.ly/31VBc07\\
\bottomrule
\end{tabular}
\caption{Summary of proposed datasets}
    \label{tab:tab_2}
\end{table}

\subsection{Cooperative patent classification (CPC)-based heuristic approach for undersampling}

As the retrieved datasets are extremely imbalanced, a model generated from these datasets would result in deficient classification performance. To handle this problem, we organize new datasets using an undersampling approach. In general, to extract a valid patent, patent experts use CPC or IPC to eliminate unrelated patents in the first step of patent landscaping. Owing to the patent characteristics, we use the CPC information to create undersampled datasets. First, we split the valid patents into a training set, validation set, and test set with a split ratio of 6, 2, and 2, respectively. Next, negative samples (i.e., not valid patents) are extracted from the retrieved search results.


We designate the negative samples from the valid patents as those not containing important CPCs. Important CPCs appear at 0.5\% or more in the valid patents for each technology area, and the emergence ratio of the CPCs in the valid patent set is more than 50 times as compared with the CPC emergence ratio in the entire USPTO patent database. This method is a reverse approach to \cite{Abood2018-fd}’s method for increasing the number of patents involved. The experiment determined the 0.5\% ratio as the minimum rate at which valid patents are not excluded. The number of important CPCs for the undersampled dataset is provided in Table \ref{tab:number_of_cpcs}. The sampled datasets are shown in Table ~\ref{tab:tab_3}.

\begin{table}[]
    \centering
    \begin{tabular}{c|c|c}
    \toprule
    Dataset name &  \# of CPCs in valid patent set & \# of important CPCs \\
    \midrule
    MPUART & 1081 & 147\\
    1MWDFS & 2543 & 145\\
    MRRG & 611 & 217\\
    GOCS & 1269 & 179\\
    \bottomrule
    \end{tabular}
    \caption{Number of important cooperative patent classifications (CPCs) in valid patents}
    \label{tab:number_of_cpcs}
\end{table}

\begin{table}[h]
\centering
\begin{tabular}{crrrr} 
\toprule
Dataset name & \# of train & \# of validation & \# of test & \# of positive\\
\midrule
MPUART & 50,280 & 10,094 & 10,094 & 280:94:94\\
1MWDFS & 50,556 & 10,185 & 10,186 & 556:185:186\\
MRRG & 50,135 & 10,045 & 10,045 & 135:45:45\\
GOCS & 50,391 & 10,131 & 10,131 & 391:131:131\\
\bottomrule
\end{tabular}
\caption{Summary of sampled datasets}
    \label{tab:tab_3}
\end{table}

\section{Deep patent landscaping model}
\subsection{Model overview}



\begin{figure*}[ht]
    \centering
    \includegraphics[width=\textwidth]{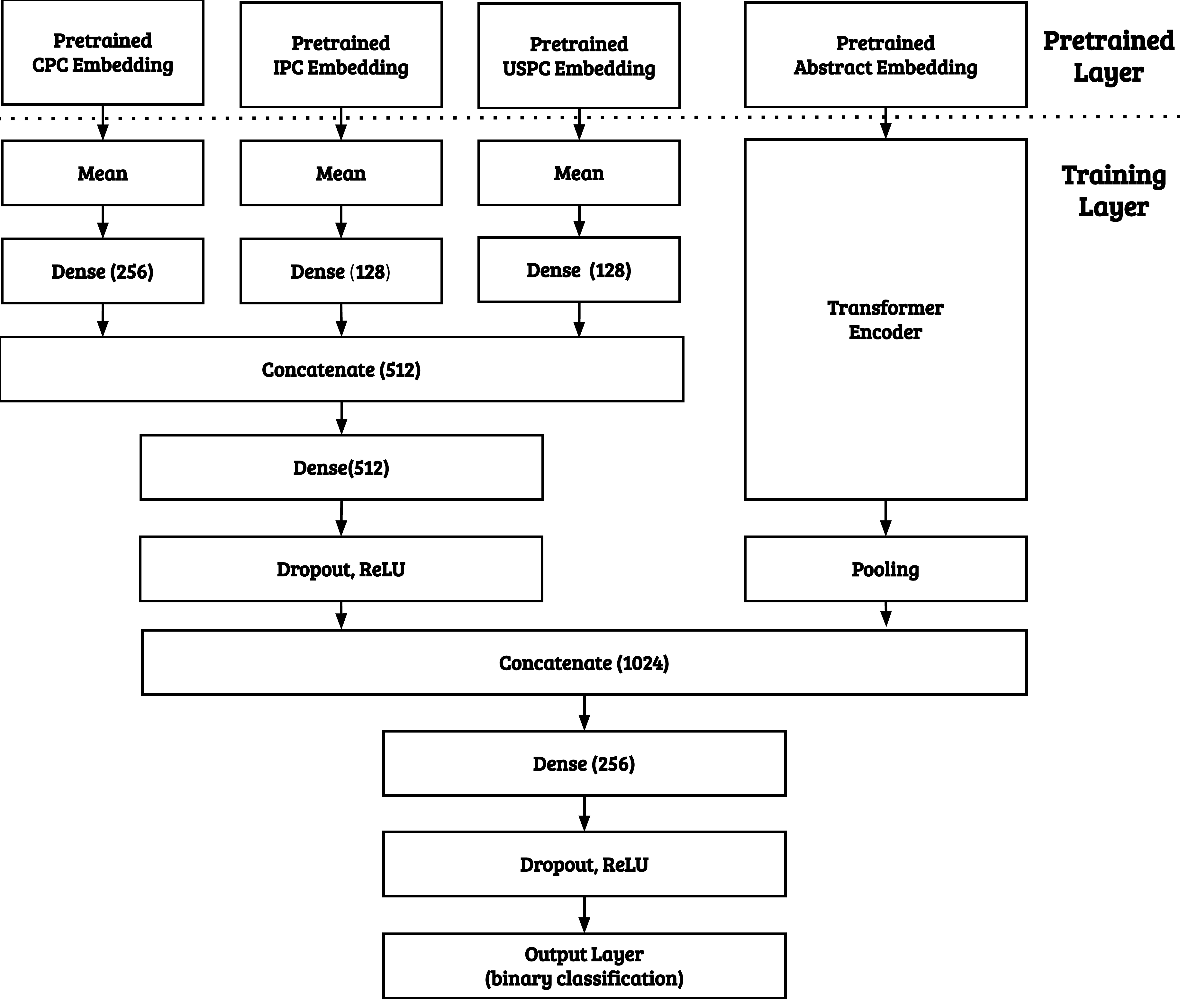}
    \caption{Architecture of a deep patent landscaping model}
    \label{fig:oveall_model}
\end{figure*}

Our proposed deep patent landscaping model is composed of two parts, as shown in Figure \ref{fig:oveall_model}: a transformer encoder(\cite{NIPS2017_7181}) and a graph embedding process using a diffusion graph called Diff2Vec(\cite{10.1007/978-3-319-73198-8_9}). The model contains a concatenation layer of embedding vectors and stacked neural network layers to classify valid patents. In that regard, a patent is a scientific document that contains textual data and metadata (i.e., fields with bibliometric information). We converted the base features of these patents into embedding spaces by considering the characteristics of each feature. Next, we trained a neural network.

\subsection{Base features}
To build a valid patent classifier, appropriate features must be selected from a patent document. Patents have a variety of features. Text data and metadata are two representative sets of features that can be used for a classification model.

Text data includes the title, abstract, description of the invention, and claims. The description of a patent is a long description of the invention, and the claims are a description of the legal rights of the invention. They are rather complicated and contain overly detailed explanations. Thus, the title and abstract, which are more general descriptions for the invention of a patent, are generally used in patent classification models(\cite{ZHANG20161108,CHEN201763,li2018deeppatent,shalaby2018lstm}).


The metadata contain a technology classification code, assignee, inventor, citations, and so on. Because the information regarding inventors and assignees is extensive, and the names may be incorrect or ambiguous, they are not suitable features for the classification model. There is also a problem that the elements of the features increase as new patents continue to issue. Therefore, technology classification codes have been continuously used in research on the patent classification. IPC and CPC are typical technology classification codes that are used in patent offices worldwide(\cite{CHEN2011309,Benson2015,doi:10.1002/asi.23664,WU2016305,PARK2017170,SUOMINEN2017131}). Countries also have their own national classification codes, such as the U.S. Patent Classification (USPC) in the US and F-term in Japan. As this research targets the USPTO dataset, we use IPC, CPC, and USPC as the basic elements for metadata. 


In summary, we use the abstract for text features and IPC, CPC, and USPC codes for metadata. To train on the features of the patents, we encode the features according to their characteristics.

\subsection{Diff2Vec for metadata embeddings}

\begin{figure*}[ht]
    \centering
    \includegraphics[width=\textwidth]{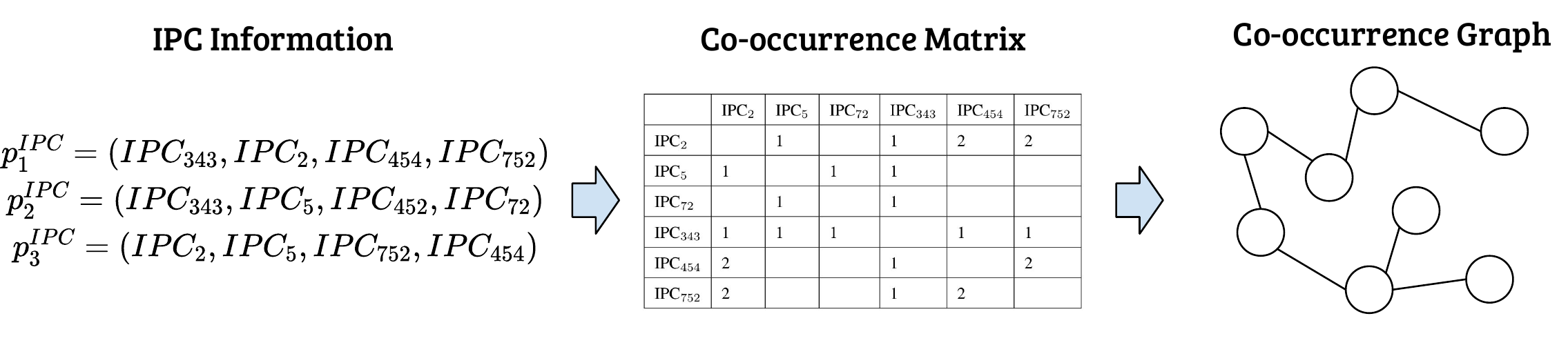}
    \caption{Transforming a technology code into a co-occurrence graph}
    \label{fig:fig_patent_landscaping_process}
\end{figure*}

We build embeddings of the technology codes, i.e., the metadata, to use them as input sources for the proposed model. 
The metadata (IPC, CPC, and USPC) are represented as a technology code information, as shown in Table \ref{tab:tab_4}.
Each technology classification code has over approximately 70,000 technology classification numbers.  Let $P=\{p_1, p_2, ... , p_n\}$ be a set of patent documents, where $n$ is the total number of patents in $P$. One document contains one or more technical codes, and we define three sets $IPC$, $CPC$, and $USPC$. Each set has their own classificaion codes. So, let $IPC=\{ipc_1, ipc_2, ..., ipc_{m^{ipc}}\}$, $CPC=\{cpc_1, cpc_2, ..., cpc_{m^{cpc}}\}$, and $USPC=\{uspc_1, uspc_2, ..., uspc_{m^{uspc}}\}$ be the sets of  $IPC$, $CPC$, and $USPC$ respectively. We define $m^{x}$ as the total number of classification codes in $IPC$, $CPC$, and $USPC$. One patent document can have  mutltiple classification codes. For example, if ${p_{32}}$ has $ipc_5$, $ipc_{102}$, and $ipc_{764}$, then we use $p_{32}^{IPC}= \{ipc_5, ipc_{102}, ipc_{764}\}$ to describe it. When several technology codes simultaneously appear in a single patent, we reflect this in a co-occurrence matrix. Next, we construct a graph. The transformation process for building the co-occurrence graph is shown in Figure \ref{fig:fig_patent_landscaping_process}.

\begin{table}[h]
\centering
\resizebox{\textwidth}{!}{
\begin{tabular}{c|c|c} 
\toprule
Code & Full name & examples\\
\midrule
IPC & International Patent Classification & E21B33/129, E21B43/11, E21B34/06\\
USPC & United States Patent Classification & 362/225., 362/230., 315/294.\\
CPC & Cooperative Patent Classification & Y02E40/642, H01L39/2419, Y10T29/49014\\
\bottomrule
\end{tabular}
}
\caption{Full names of classification codes}
    \label{tab:tab_4}
\end{table}



\begin{figure*}[ht]
    \centering
    \includegraphics[width=\textwidth]{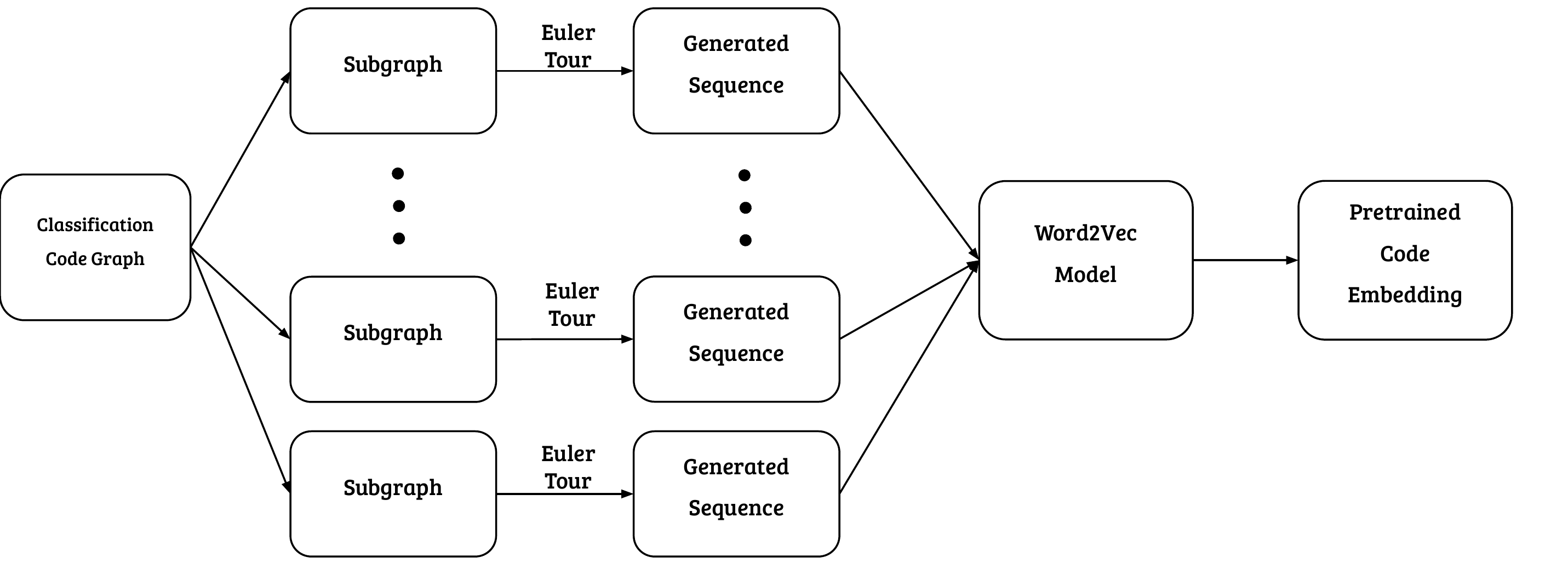}
    \caption{Pretraining of metadata graph embeddings}
    \label{fig:graph-embedding-process}
\end{figure*}

After transforming the metadata into a graph representation, we adopt the Diff2Vec method for the graph representation, to place it into the proposed neural network model. Diff2Vec is a graph embedding method based on Word2Vec(\cite{NIPS2013_5021}). It uses a diffusion process to extract a neighbor node’s subgraph, called a diffusion graph. The subgraph is formed by being diffused by neighboring nodes that are randomly selected based on one node in the subgraph. Then, a Euler tour is applied to the diffusion graph to generate a sequence. The sequences generated by the Euler tour are used to train the Word2Vec layer. We set the length of the diffusion at 40, and the number of diffusions per node at 10. According to experiments, Diff2Vec scales better as the graph’s density increases, and the embedding preserves graph distances with high accuracy. In our model’s architecture, we used a pretrained Diff2Vec for the embedding layer of three classification codes. We averaged the embedding values of each code to combine the graph information for one patent. Then, we used a dense layer for processing the averaged graph information. We process the CPCs to 256, twice the Diff2Vec embedding size, and the other codes to 128. This is because CPC is the most granular classification code; thus, we wanted to use more information regarding CPC than other codes. The detailed pretraining process for metadata is shown in Figure \ref{fig:graph-embedding-process}.

\subsection{Transformer architecture for text data}

Another core building block of our model is the transformer layer for the text data. To handle text data, we extracted abstracts of each patent, divided paragraphs via tokens, and built embeddings of the tokens using Word2Vec(\cite{NIPS2013_5021}). When we tokenized the abstract text, the tag [CLS] was inserted at the beginning of the first sentence, and the tag [SEP] was inserted at the end of the sentence. Then, we transmitted the embeddings to the transformer encoder (\cite{NIPS2017_7181}) to learn the latent space for the patent abstract paragraph. We stacked the encoder layer 6 times. We also used multi-head self-attention and scaled dot-product attention, without modifying the transformer encoder. We set the number of heads of the multi-head self-attention at 8. We set the sequence length to 128, and the hidden size was 512. 

\subsection{Training and inference phrase}


Finally, we add abstraction embedding vectors from the metadata and text data by concatenating both, and we input them into a simple multilayer perceptron (MLP). To concatenate the output of the transformer with the classification code embedding vectors, we adopted a squeeze technique from the “Bidirectional Encoder Representations from Transformers” (BERT, \cite{devlin-etal-2019-bert}) and converted the matrix  to a vector (embedding size) based on the [CLS] tag. To classify whether a target patent is a valid patent or not, we use binary cross-entropy in the last layer.

\section{Experiments}

\subsection{Dataset}
We measured the performance of the proposed model for the classification of valid patents in the four KISTA datasets. More than half of the datasets had over one million documents. In this case, those large datasets may contain search formula keywords but also contain noisy patents (which are out of the domain). Moreover, extracting embeddings from those datasets and using them for model training requires significant computing resources. Thus, we use high-frequency CPC codes for heuristic sampling to filter the noisy data.

\subsection{Hyperparameter settings}

Six encoder layers were stacked in the transformer, and the number of attention heads was eight. Another model consisted of 12 encoder layers and four attention heads. The number of learning epochs, batch size, optimizer, learning rate, and epsilon were set as follows: 20, 64, Adam optimizer(\cite{2014arXiv1412.6980K}), 0.0001, and $1e-8$, respectively. We set the sequence length, i.e., the maximum length of the input sentence, to 128, and padded it to 0 if it was shorter than 128. As a result, 512-dimensional embedding vectors were extracted for each word.

\subsection{Evaluation metric}

We used the average precision and F1-score as evaluation metrics, which are commonly used in binary classification problems for imbalanced datasets. We compare the following models: APL(\cite{Abood2018-fd})\footnote{We modified APL's code to be worked on our dataset.}, Word2Vec, and Diff2Vec-based classifiers5.

\section{Results of experiments }
\subsection{Overall results}

For each patent, our model considers two sets of features: metadata and text data. We experiment with our proposed model to determine how each feature affects classification performance. For the metadata, we identified how CPC, IPC, and USPC affect the performance. IPC is an internationally unified patent classification system with five hierarchies and approximately 70,000 codes. USPC is a US patent classification system based on claims, with approximately 150,000 codes. CPC is the latest patent classification system, which reflects new classifications according to technological developments. CPC is a more detailed classification system than IPC. It was developed based on the European Classification System and USPC, and it includes approximately 260,000 codes. We identify how the transformer configuration affects the text data, from the perspective of classification performance. We compare the classification performance of our model with APL, i.e., the latest patent landscaping deep learning model. The experimental results show that our model outperforms all other models. Moreover, our model performs well even when classifying using only classification codes. The overall results are shown in Table\ref{tab:overall_results}.

\begin{table*}[htbp]
\resizebox{\textwidth}{!}{
    \centering
    \begin{tabular}{c|c|c|c|c|c|c|c|c}
    \toprule
    \multirow{2}*{Dataset} & \multicolumn{2}{c|}{TRF+DIFF} & \multicolumn{2}{c|}{TRF} & \multicolumn{2}{c|}{DIFF} & \multicolumn{2}{c}{APL} \\ \cline{2-9}
        & AP & F1 & AP & F1 & AP & F1 & AP & F1 \\
    \midrule
    MPUART & \textbf{0.6552} & \textbf{0.8025} & 0.4746 & 0.6684 & 0.6045 & 0.7711 & 0.3028 & 0.5340 \\ 
    1MWDFS & \textbf{0.566} & \textbf{0.7438} & 0.4527 & 0.6564 & 0.5429 & 0.7285 & 0.4155 & 0.6055 \\ 
    MRRG & \textbf{0.6871} & \textbf{0.823} & 0.4960 & 0.6988 & 0.6792 & 0.8208 & 0.2065 & 0.4086 \\ 
    GOCS & \textbf{0.4286} & \textbf{0.6467} & 0.3742 & 0.5966 & 0.3825 & 0.6019 & 0.3277 & 0.5424 \\ 
    \bottomrule
    \end{tabular}
}
    \caption{Average precision and F1-scores of the baseline and the proposed model}
    \label{tab:overall_results}
\end{table*}

\subsection{Effects of technology code metadata}

As shown in Table \ref{tab:overall_results_for_code}, we conducted experiments for each code to analyze the effects of each code. As a result, CPC, the most subdivided classification, showed the highest classification performance. However, the performance of USPC was slightly higher than that of CPC for geostationary orbit complex satellite (GOCS) data. Therefore, we performed a quantitative analysis to investigate it. For a fair comparison, the dimensionality of the density layer (after the graph embedding layer) is 128 for all classification codes.

\begin{table*}[htbp]
    \centering
    \resizebox{\textwidth}{!}{
    \begin{tabular}{c|c|c|c|c|c|c|c|c}
    \toprule
    \multirow{2}*{Dataset} & \multicolumn{2}{c|}{TRF+DIFF} & \multicolumn{2}{c|}{text+cpc} & \multicolumn{2}{c|}{text+ipc} & \multicolumn{2}{c}{text+uspc} \\ \cline{2-9}
        & AP & F1 & AP & F1 & AP & F1 & AP & F1 \\
    \midrule
    MPUART & \textbf{0.6552} & \textbf{0.8025} & \textcolor{red}{0.6321} & \textcolor{red}{0.7835} & 0.586 & 0.7606 & 0.5372 & 0.7227 \\ 
    1MWDFS & \textbf{0.566} & \textbf{0.7438} & \textcolor{red}{0.5384} & \textcolor{red}{0.7069} & 0.4902 & 0.6883 & 0.4669 & 0.6776 \\ 
    MRRG & \textbf{0.6871} & \textbf{0.823} & \textcolor{red}{0.6634} & \textcolor{red}{0.8069} & 0.5067 & 0.7059 & 0.6195 & 0.7814 \\ 
    GOCS & \textbf{0.4286} & \textbf{0.6467} & 0.4071 & 0.6301 & 0.3922 & 0.6151 & \textcolor{red}{0.4140} & \textcolor{red}{0.6347} \\ 
    \bottomrule
    \end{tabular}
    }
    \caption{Assessing influence by code}
    \label{tab:overall_results_for_code}
\end{table*}

\subsection{Effects of text data}
We experimented with different sizes of transformers and several text-embedding methods. Our proposed model shows high performance for most datasets. However, the micro-radar rain gauge (MRRG) dataset provides better performance with different hyperparameters of the transformer configuration. The MRRG dataset had significantly worse classification performance than other datasets. For this reason, we believe that organizing the transformer structure for text more deeply than using codes alone shows better performance. In other words, if the number of valid patents is small, there is more reliance on the text than on technology codes. Moreover, we found that the MRRG dataset’s average sequence length was the shortest; therefore, it could achieve high performance with only four attention heads. In addition, the overall performance difference was not significant when using other text embedding techniques. However, Doc2Vec’s performance was better than the other embedding techniques.


\begin{table*}[htbp]
    \centering
    \resizebox{\textwidth}{!}{
    \begin{tabular}{c|c|c|c|c|c|c|c|c|c|c}
    \toprule
    \multirow{2}*{Dataset} & \multicolumn{2}{c|}{TRF(6,8)+DIFF} & \multicolumn{2}{c|}{TRF(12,4)+DIFF} & \multicolumn{2}{c|}{Word2Vec+DIFF} & \multicolumn{2}{c|}{Doc2Vec+DIFF} & \multicolumn{2}{c}{Fasttext+DIFF} \\ \cline{2-11}
        & AP & F1 & AP & F1 & AP & F1 & AP & F1 & AP & F1\\
    \midrule
    MPUART & \textbf{0.6552} & \textbf{0.8025} & 0.6208 & 0.7810 & 0.6183 & 0.7739 & 0.65 & 0.7975 & 0.6165 & 0.7748 \\ 
    1MWDFS & \textbf{0.566} & \textbf{0.7438} & 0.5667 & 0.7404 & 0.5279 & 0.7123 & 0.556 & 0.7312 & 0.5371 & 0.7083 \\ 
    MRRG & 0.6871 & 0.823 & \textbf{0.7384} & \textbf{0.8426} & 0.6414 & 0.7895 & 0.7020 & 0.8289 & 0.6835 & 0.8212\\ 
    GOCS & \textbf{0.4286} & \textbf{0.6467} & 0.3845 & 0.6027 & 0.3603 & 0.5918 & 0.3915 & 0.6148 & 0.3367 & 0.5556\\ 
    \bottomrule
    \end{tabular}
    }
    \caption{Comparison with the embedding models}
    \label{tab:overall_results}
\end{table*}

\subsection{Lessons learned from the experiments}
The following lessons were learned from the experiment results of the patent classification model. 

\begin{itemize}
\item Patent documents comprise large amounts of scholarly data containing metadata and text data. It was found that classifying patent documents using both sets of features is important for providing better classification performance, as contrasted with using an individual feature alone.
\item Technology codes play a vital role in patent document classification. This may be because technology codes are often used as the primary criterion for classification when experts conduct patent classifications.
\item The important technology classification codes may vary depending on the characteristics of the dataset. In general, however, CPCs, which are more detailed technology codes, guarantee better results in classification performance.
\item Depending on the dataset, other technology codes may become more important. The number of technology codes that a valid patent has in that dataset is an important feature for patent classification. For example, in the case of the GOCS dataset, USPC has a slightly higher impact on classification performance, as the number of USPC codes in the valid patents is proportionally much higher than the CPCs.
\item In the case of datasets with a more extreme imbalance, it may be helpful to study the transformer more deeply than simply the effects of the technology codes. When the number of CPC codes of valid patents is reduced, the model learns the classification pattern from text data.
\item As in any other text classification model, high performance is shown for patent documents when a transformer architecture is used. However, given the efficiency of the model, Doc2Vec can also be a good alternative for text data.
\end{itemize}


\section{Conclusion}
In this paper, we proposed a deep patent landscaping model that addresses the classification problem in patent landscaping using a transformer and Diff2Vec structures. Our study contributes to the research on patent landscaping in three aspects. First, we introduced a new benchmarking dataset for automated patent landscaping and provided a practical study for automated patent landscaping. Second, our model showed a high overall classification performance in patent landscaping, as compared to existing models. Finally, we experimentally analyzed how the technical codes and text data affect models in patent classification. We believe this research will help to reduce the repetitive patent analysis tasks required of practitioners.

Further research is required on patent classification. There are various metadata in patent documents, such as assignees, inventors, and citations. One could identify whether including these features would improve classification performance. Additionally, different datasets require different types of classification models. We need to develop models that fit different datasets. It is expected that this can be addressed through research on meta-learning and AutoML, which are the current topics in the field of deep learning.

\section{Acknowledgement}
This work was supported by the National Research Foundation of Korea (NRF) grant and funded by the Korean government (No. NRF-2015R1C1A1A01056185 and No. NRF-2018R1D1A1B07045825). We really appreciate Ph.D. Min and Ph.D. Kim, living in southern area of Gyeonggi-do in Korea. They gave us a lot of inspiration and courage to write this paper.

\newpage

\appendix
\section{BigQuery Search Query for Patent Datasets}
\begin{center}
\begin{longtable}{| c | p{.80\textwidth} |} 
    \hline
    Dataset Name & Query \\ \hline
    \multirow{27}*{MPUART} & {\scriptsize (((REGEXP\_CONTAINS(description.text, " virtual\%") or REGEXP\_CONTAINS(description.text, " augment\%") or REGEXP\_CONTAINS(description.text, "mixed\%")) or (REGEXP\_CONTAINS(description.text, " real\%") or REGEXP\_CONTAINS(description.text, " environment\%") or REGEXP\_CONTAINS(description.text, " space "))) or (REGEXP\_CONTAINS(description.text, " augment\%") and REGEXP\_CONTAINS(description.text, " real\%"))) and (((REGEXP\_CONTAINS(description.text, " offshore\%") or REGEXP\_CONTAINS(description.text, " off-shore\%") or REGEXP\_CONTAINS(description.text, " ocean ")) or (REGEXP\_CONTAINS(description.text, " plant\%") or REGEXP\_CONTAINS(description.text, " platform\%"))) or REGEXP\_CONTAINS(description.text, " ship\%") or REGEXP\_CONTAINS(description.text, " dock\%") or REGEXP\_CONTAINS(description.text, " carrier ") or REGEXP\_CONTAINS(description.text, " vessel ") or REGEXP\_CONTAINS(description.text, " marine ") or REGEXP\_CONTAINS(description.text, " boat\%") or REGEXP\_CONTAINS(description.text, " drillship ") or (REGEXP\_CONTAINS(description.text, " drill ") or REGEXP\_CONTAINS(description.text, " ship ")) or REGEXP\_CONTAINS(description.text, " FPSO ") or (REGEXP\_CONTAINS(description.text, " float\%") or (REGEXP\_CONTAINS(description.text, " product\%") or REGEXP\_CONTAINS(description.text, " storag\%"))) or REGEXP\_CONTAINS(description.text, " FPU ") or REGEXP\_CONTAINS(description.text, " LNG ") or REGEXP\_CONTAINS(description.text, " FSRU ") or REGEXP\_CONTAINS(description.text, " OSV ") or REGEXP\_CONTAINS(description.text, " aero\%") or REGEXP\_CONTAINS(description.text, " airplane ") or REGEXP\_CONTAINS(description.text, " aircraft ") or REGEXP\_CONTAINS(description.text, " construction ") or (REGEXP\_CONTAINS(description.text, " civil ") or REGEXP\_CONTAINS(description.text, " engineer\%")) or REGEXP\_CONTAINS(description.text, " bridge ") or REGEXP\_CONTAINS(description.text, " building ") or REGEXP\_CONTAINS(description.text, " vehicle ") or REGEXP\_CONTAINS(description.text, " vehicular ") or REGEXP\_CONTAINS(description.text, " automotive ") or REGEXP\_CONTAINS(description.text, " automobile "))}\\ \hline
    \multirow{15}*{1MWDFS} & {\scriptsize(((REGEXP\_CONTAINS(description.text, " inducti\%") or REGEXP\_CONTAINS(description.text, " heating ")) or (REGEXP\_CONTAINS(description.text, " induction ") or REGEXP\_CONTAINS(description.text, " hardening ")) or (REGEXP\_CONTAINS(description.text, " contour ") or REGEXP\_CONTAINS(description.text, " hardening ")) or (REGEXP\_CONTAINS(description.text, " surface ") or REGEXP\_CONTAINS(description.text, " hardening "))) and (REGEXP\_CONTAINS(description.text, " dual-frequency ") or REGEXP\_CONTAINS(description.text, " multi-frequency ") or ((REGEXP\_CONTAINS(description.text, " dual ") or REGEXP\_CONTAINS(description.text, " multi ")) or REGEXP\_CONTAINS(description.text, " frequency ")) or (REGEXP\_CONTAINS(description.text, " frequency ") or (REGEXP\_CONTAINS(description.text, " selectable ") or REGEXP\_CONTAINS(description.text, " variable "))))) or ((REGEXP\_CONTAINS(description.text, " Inducti\%") or REGEXP\_CONTAINS(description.text, " heating ")) and ((REGEXP\_CONTAINS(description.text, " contour ") or REGEXP\_CONTAINS(description.text, " hardening ")) or (REGEXP\_CONTAINS(description.text, " surface ") or REGEXP\_CONTAINS(description.text, " hardening "))))} \\ \hline
    \multirow{22}*{MRRG} & {\scriptsize ((REGEXP\_CONTAINS(description.text, " precipitat ") or REGEXP\_CONTAINS(description.text, " rain ") or REGEXP\_CONTAINS(description.text, " snow ") or REGEXP\_CONTAINS(description.text, " weather ") or REGEXP\_CONTAINS(description.text, " climate ") or REGEXP\_CONTAINS(description.text, " meteor ") or REGEXP\_CONTAINS(description.text, " downpour ") or REGEXP\_CONTAINS(description.text, " cloudburst ") or REGEXP\_CONTAINS(description.text, " deluge ") or REGEXP\_CONTAINS(description.text, " flood ") or REGEXP\_CONTAINS(description.text, " disaster ") or (REGEXP\_CONTAINS(description.text, " wind ") or (REGEXP\_CONTAINS(description.text, " field ") or REGEXP\_CONTAINS(description.text, " speed ") or REGEXP\_CONTAINS(description.text, " velocit ") or REGEXP\_CONTAINS(description.text, " direction "))) or REGEXP\_CONTAINS(description.text, " storm ") or REGEXP\_CONTAINS(description.text, " hurricane ")) and ((REGEXP\_CONTAINS(description.text, " radio ") or (REGEXP\_CONTAINS(description.text, " wave ") or REGEXP\_CONTAINS(description.text, " signal ") or REGEXP\_CONTAINS(description.text, " frequency "))) or ((REGEXP\_CONTAINS(description.text, " electr ") or REGEXP\_CONTAINS(description.text, " micro ")) or REGEXP\_CONTAINS(description.text, " wave ")) or REGEXP\_CONTAINS(description.text, " beam ")) and (REGEXP\_CONTAINS(description.text, " verif ") or REGEXP\_CONTAINS(description.text, " check ") or REGEXP\_CONTAINS(description.text, " invest ") or REGEXP\_CONTAINS(description.text, " experiment ") or REGEXP\_CONTAINS(description.text, " test ") or REGEXP\_CONTAINS(description.text, " simulat ")))} \\ \hline
    \multirow{6}*{GOCS} & {\scriptsize ((REGEXP\_CONTAINS(description.text, " satellite ")) and (REGEXP\_CONTAINS(description.text, " band ") or REGEXP\_CONTAINS(description.text, " illumination ") or REGEXP\_CONTAINS(description.text, " illuminance ")) and (REGEXP\_CONTAINS(description.text, " merge ") or REGEXP\_CONTAINS(description.text, " merging ") or REGEXP\_CONTAINS(description.text, " fusion ") or REGEXP\_CONTAINS(description.text, " mosaic ")))} \\ \hline
\end{longtable}
\end{center}

\bibliography{main}

\end{document}